# Value of structural health information in partially observable stochastic environments

C.P. Andriotis[1], K.G. Papakonstantinou[1], E.N. Chatzi[2]

[1]Department of Civil & Environmental Engineering, The Pennsylvania State University, University Park, PA, USA
[2]Department of Civil, Environmental and Geomatic Engineering, ETH Zürich, Zürich, Switzerland

**Abstract**

Efficient integration of uncertain observations with decision-making optimization is key for prescribing informed intervention actions, able to preserve structural safety of deteriorating engineering systems. To this end, it is necessary that scheduling of inspection and monitoring strategies be objectively performed on the basis of their expected value-based gains that, among others, reflect quantitative metrics such as the Value of Information (VoI) and the Value of Structural Health Monitoring (VoSHM). In this work, we introduce and study the theoretical and computational foundations of the above metrics within the context of Partially Observable Markov Decision Processes (POMDPs), thus alluding to a broad class of decision-making problems of partially observable stochastic deteriorating environments that can be modeled as POMDPs. Step-wise and life-cycle VoI and VoSHM definitions are devised and their bounds are analyzed as per the properties stemming from the Bellman equation and the resulting optimal value function. It is shown that a POMDP policy inherently leverages the notion of VoI to guide observational actions in an optimal way at every decision step, and that the permanent or intermittent information provided by SHM or inspection visits, respectively, can only improve the cost of this policy in the long-term, something that is not necessarily true under locally optimal policies, typically adopted in decision-making of structures and infrastructure. POMDP solutions are derived based on point-based value iteration methods, and the various definitions are quantified in stationary and non-stationary deteriorating environments, with both infinite and finite planning horizons, featuring single- or multi-component engineering systems.

## 1. Introduction

The development of new monitoring technologies, data acquisition techniques and information processing methodologies further encourages the use of Structural Health Monitoring (SHM) in supporting management of critical infrastructure and deteriorating systems [1, 2]. These new possibilities come with relevant questions related to the actual value and necessity of increased quality measurements or continuous SHM information in facilitating optimal actions. SHM is defined as the development of online and automated damage detection capabilities for all types of aerospace, civil and mechanical infrastructure [3], and these SHM aspects distinguish it from traditional non-destructive evaluation or inspection-based approaches, often conducted in a targeted and periodic manner. Along these lines, SHM frameworks seek to determine appropriate mappings from raw response measurements to condition and performance indicators, which can, subsequently, support decision-making towards cost-effective intervention and maintenance actions that increase safety and mitigate risks [4]. Quantifying the overall information gains of SHM systems is thus a multi-stage process. First, the SHM data have to be collected and processed, and relevant features need to be selected (instrumentation and data preparation stage). These features can be then mapped into proper descriptive and predictive models that can efficiently infer quantities of interest over future time steps (learning stage). This stage often involves parameter identification techniques, statistical fitting and error estimation approaches, and can be performed within the concepts of a great span of uncertainty quantification and machine learning methods. Lastly, data and models are integrated with optimization and decision analysis frameworks and, hence, relevant long-term and/or global response metrics can be controlled and quantified (decision stage).

This work focuses on the final decision stage, with the estimated values of performance/condition indicators, as these are extracted from the raw data in post-processing, being treated as uncertain environment *observations*. Naturally, the uncertainty in observations stems from the uncertainty in the estimation of those indicators, as this is induced by statistical errors or variability in the outcomes of the measuring instruments. More specifically, the decision-making stage pertains to the type and sequence of actions that are selected in order to optimize an overarching predefined life-cycle objective. As such, when the objective is to maximize long-term safety and resilience, and to effectuate preventive maintenance actions, SHM typically constitutes a natural choice, as it can be used to diagnose faults and even determine the root cause of the fault process, e.g. [5]. However, to what measurable extent is the acquired information able to support improved policy-planning in an engineering environment, and how can we objectively quantify the resulting gains?

Rational frameworks for decision and information analysis have long been a topic of interest beyond engineering and infrastructure management [6, 7, 8]. Within infrastructure management and planning, an important discussion in this respect is whether the benefits of the various observational strategies, e.g. SHM-aided plans, or in situ visual and specialized non-destructive evaluation inspections, can be quantified in terms of life-cycle value-based metrics, and whether these benefits are comparable to the costs associated with the use of SHM equipment (acquirement, installation, operation, maintenance costs, etc.). The question that summarizes this discussion is *how much is information worth* or, similarly, *how much is a SHM system worth investing?* [9, 10, 11]. In response, recent research efforts focus on quantifying the Value of Information (VoI) and, similarly, the Value of Structural Health Monitoring (VoSHM) within rigorous mathematical frameworks. Following the definitions in [12], VoI may be defined in pertinence to inspection and maintenance planning, and may as such be devised along the lines of pre-posterior engineering decision problems [13, 14, 15]. The concept of VoI can be utilized to (i) evaluate the amount the decision-maker is willing to pay for information prior to a single decision step of the decision process, either considering the long- or short-term information benefits, e.g. [16] or [17], respectively; or (ii) to quantify the overall gain that information may yield regarding a fixed inspection and/or monitoring policy, applied over the entire life-cycle of a system, e.g. [18]. The latter measure of VoI may be used to assess whether it is worth adopting a certain observational strategy over others, from the beginning up to the end of the system's operational life, and this is the approach followed in this paper. Similarly, within the context of SHM, VoI may be quantified as the difference between the expected cost of maintaining the system in absence of SHM information, and the cost given availability of monitoring information [10, 19, 9, 20]. Along these lines, within the context of Partially Observable Markov Decision Processes (POMDPs), VoI analysis and quantification approaches have been also developed in [21, 22, 23]. VoSHM is herein defined as a more specialized definition of VoI, describing relative costs between intermittent/optional observational schemes, e.g. periodic or non-periodic inspection visits, and SHM-aided plans, where the flow of observations is typically continuous [24].

As already mentioned, the VoI and VoSHM metrics may be quantified as per their impact in rendering infrastructure management more effective. One approach to quantifying these metrics is to formulate an optimization problem which seeks to determine optimal sequences of actions (policies) and their respective life-cycle costs for different observational scenarios. Key to the success of such optimization formulations is (i) incorporation of environment stochasticity, (ii) long-term optimality of decisions, and (iii) integration of dynamic, real-time, noisy observations. Numerous formulations exist in the literature dealing with the issue of decision-making for optimal management of infrastructure [25]. Typically, the objective function pertains to various life-cycle conditions, reliability, risk and cost measures, which are sought to be optimized by the decision-maker. These, as well as the employed optimization approaches and environment simulators may vary depending on the system specifications. Dynamic Bayesian networks are utilized in [26], to determine the underlying structural deterioration process. Based on the established dependencies, the cumulative life-cycle cost is evaluated and the policy space can be subsequently searched through optimization heuristics [16, 27, 28], genetic algorithms [29], or other relevant optimization solvers. POMDPs are also built within dynamic Bayesian network premises, so the two approaches can be seen as equivalent in terms of how the environments are simulated, however, their adopted optimization approaches and capabilities are completely different [30]. Renewal processes can also be utilized in this regard, with various extensions and refined formulations to account for multi-threshold and multi-level action plans, or even integrated resilience considerations [31, 32, 33]. Within this context, direct search of the discretized decision variable space can be conducted to determine the best strategy, or even analytical and gradient-based approaches, if applicable. Multi-criteria objectives have been also examined in [34, 35], to account for a diversified quantification of risk, in its socioeconomic and environmental constituents, or to seek optimized values of competitive cost and performance indicators [36]. In such cases, optimization can be efficiently conducted using heuristics, such as genetic algorithms.

In this work, the inspection and maintenance optimization problem is addressed within the framework of POMDPs. Primarily developed in the field of robotics over the past years for stochastic optimal control in partially observable dynamic environments, POMDPs provide a well-suited mathematical framework for sequential decision-making, with sound life-cycle optimality guarantees and convergence properties [37], which can be conveniently lent to the class of structural inspection and maintenance problems [38]. POMDPs extend Markov Decision Processes (MDP) to partially observable environments, where the decision-maker/agent seeks to optimize a policy maximizing the collected rewards over time (or minimizing costs), without knowing the exact state of the system. In [39, 40], POMDPs are adopted for decision-making for highway-pavements. The use of POMDPs has been also applied in [41, 42] for bridge inspection planning, whereas point-based solutions for stochastic deteriorating systems using POMDPs have been presented in [43, 44, 45, 46]. A continuous formulation for problems described by linear and/or nonlinear transition functions is presented in [47], whereas specialized cases of mixed observability are also presented in [48]. Exploiting VoI, POMDPs can further be extended to tackle inspection and maintenance problems at the system level, as in [22]. Recent frameworks within the premises of deep reinforcement learning, particularly efficient in addressing the curse of dimensionality and model unavailability issues in large-scale POMDP system applications, are developed in [49, 50, 51]. Regardless of the adopted numerical solution scheme (i.e., alpha-vector value iteration, point-based value iteration, reinforcement learning, etc.), POMDPs are particularly favorable for decision-making formulations in infrastructure management and have been demonstrated to significantly outperform conventional fixed inspection and maintenance policies [49, 52]. This is particularly true in the presence of discrete or discretized spaces, where exhaustive evaluation and search of policy subspaces, evolutionary approaches, or gradient methods may be impractical or ineffective, if at all applicable. Moreover, regarding the step-wise definition of VoI, i.e., the amount the decision-maker is willing to pay prior to each decision, as also rigorously discussed in Section 3.1, it is worth noting that POMDPs inherently and straightforwardly leverage VoI, if observation actions are introduced as separate decision variables, since the Bellman equation of optimality that POMDP solutions satisfy, minimizes the future cumulative cost at each decision step after considering all possible alternative observational choices.

The developed methodology for calculating the VoI and VoSHM in this work is primarily aimed at assessing their life-cycle aspects, thus targeting decision-making for the selection of long-term observational plans among various alternatives. Detailed definitions of the above value metrics are devised and discussed, their theoretical mathematical properties are analyzed, and the underlying steps for their computation are demonstrated in numerical experiments of deteriorating engineering systems operating in partially observable stochastic environments. It is formally proven that the two metrics are non-negative for the optimal POMDP policy, thus the additional information that inspections and SHM provide can only improve decisions in the long-run. Moreover, a step-wise VoI metric is defined, which allows for a convenient reformulation of the Bellman equation, showcasing that POMDPs utilize the notion of VoI at every decision step as the criterion to optimally choose observational actions. Quantification of the above metrics is based on solutions derived through POMDP formulations for the inspection and maintenance optimization problem, however, the applicability of the method is not particular to POMDPs and can be easily adjusted to the needs and outcomes of other optimization schemes. An infinite horizon three-component POMDP system and a larger finite horizon POMDP problem, modeling a deteriorating port deck structure, are analyzed under two different inspection scenarios; one with optional inspection visits and one with continuous availability of observations, resembling a SHM system. The underlying POMDPs are solved using various point-based value iteration algorithms, which are shown to provide particularly effective solutions for the proposed framework. The described VoI and VoSHM analysis provides the respective expected gains in terms of a life-cycle metric of interest, e.g. cost, thus answering the previously posed question of how much is inspection or monitoring information eventually worth, as well as how information of increased precision can affect decisions. In particular, VoI quantifies the value that is added by the availability of observation choices over the system lifetime, whereas the VoSHM quantifies the possible benefits of adopting a monitoring system from the beginning of the planning horizon, over following a plan based on optional inspection visits.

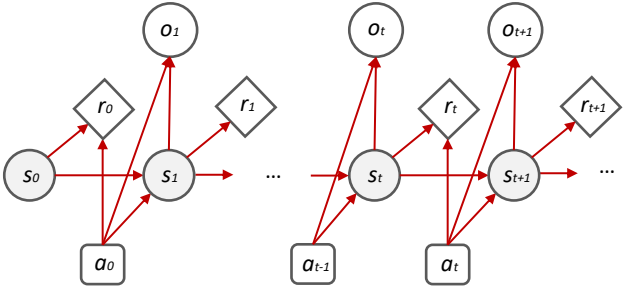

**Fig. 1**: Probabilistic graphical model of a POMDP in time (shaded nodes denote hidden states).

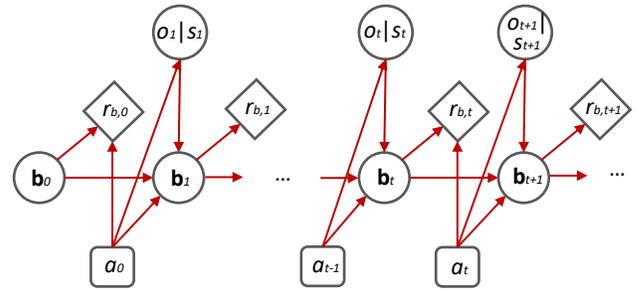

**Fig. 2**: Probabilistic graphical model of a POMDP as a belief-MDP in time (observations depend on states which are hidden).

## 2. Partially Observable Markov Decision Processes

POMDPs provide an adept framework for stochastic optimal control. They are established within the premises of dynamic programming, thus providing strong global optimality guarantees for long-term decision problems described by stochastic environment dynamics with Markovian properties, noisy observations and uncertain action outcomes. Markovian assumptions do not restrict the applicability of POMDPs in non-Markovian environments, as the latter can be properly transformed to fit Markovian assumptions through state augmentation, as discussed in [53, 43, 27]. POMDPs generalize Markov Decision Processes (MDPs) to partially observable environments, i.e., to cases where observations are unable to reveal the actual state of the system with certainty. This feature, along with their neat mathematical formulation of POMDPs, is suitable to efficiently describe inspection and maintenance planning problems in structural and generic engineering settings, where the inspection techniques and monitoring devices deployed, typically provide incomplete information about the system condition (states), which evolves according to an underlying stochastic deterioration process.

According to the POMDP problem statement, the decision-maker/agent starts at a state, $s_t = s \in S$ at every decision step, $t$, takes an action, $a_t = a \in A$, receives a reward, $r = r(s,a)$, transitions to the next state, $s_{t+1} = s'$, according to a Markovian transition probability model conditioned at the current state and action, $p(s'|s,a)$, and receives an observation, $o_{t+1} = o \in \Omega$, based on its state and action, according to the probability defined by an observation model, $p(o|s',a)$. This process is schematically depicted in Fig. 1. More formally, a POMDP is a 7-tuple $\mathcal{L} = \langle S, A, \mathbf{P}, \Omega, \mathbf{O}, \mathbf{R}, \gamma \rangle$ where $S$, $A$ and $\Omega$ are finite sets of states, actions and possible observations, respectively. $\mathbf{P}$, $\mathbf{O}$ are the 3-dimensional Markovian state transition and observation probability matrices, respectively, whereas $\mathbf{R}$ is the reward matrix, defined as:

$$\begin{aligned}\mathbf{P} &= \left[\mathbf{P}_a\right]_{a \in A} = \left[p(s'|s,a)\right]_{s,s' \in S, a \in A} \\ \mathbf{O} &= \left[\mathbf{O}_a\right]_{a \in A} = \left[p(o|s',a)\right]_{o \in \Omega, s' \in S, a \in A} \\ \mathbf{R} &= \left[\mathbf{R}_a\right]_{a \in A} = \left[r(s,a)\right]_{s \in S, a \in A}\end{aligned} \quad (1)$$

As a result of partial observability, at every decision step $t$, the agent cannot be fully aware of its state, $s_t$ (shaded nodes in Fig. 1), which may only be perceived through an observation $o_t$ that is a noisy indicator of that state [54].

Starting with an initial distribution of state $s_0$ over $S$, the objective of the agent is to determine a sequence of actions that maximizes the expected return, i.e., the expected total cumulative future reward. This is accomplished by executing an optimal policy $\pi = \pi^*$, which maps the history of actions and observations up to time $t$, to the current action $a_t$, such that:

$$\pi^* = \arg\max_{\pi} \mathbb{E}_{s_{t \geq 0}, o_{t > 0}}\left[\sum_{t=0}^{\infty} \gamma^t r(s_t, a_t) \bigg| a_t = \pi(a_0, o_1, \ldots, a_{t-1}, o_t)\right] \quad (2)$$

where $\gamma$ is the discount factor, a positive scalar less than 1, associated with the present value of future rewards. In the context of inspection and maintenance planning, rewards are typically negative quantities describing costs. It can also be noted that Eq. (2) describes an infinite horizon problem. Assumed operation over an infinite number of steps offers the advantage of not arbitrarily predefining the end of operational life. It is also not restrictive in terms of modeling, in cases where such analysis is not relevant, since finite horizon problems can be also formulated as infinite horizon ones, with proper consideration of time-related states and an introduced absorbing state at the final time step [53]. Relevant implementation aspects are also discussed in the numerical examples in Section 4.2.1.

Although the agent cannot observe the exact state with certainty as a result of partial observability, it can form a belief $b_t = b \in B$ about its state, where $b$ is a probability distribution over set, $S$, of all possible discrete states. Space $B$ is a $(|S|-1)$-dimensional simplex. The new belief $b_{t+1} = b'$, i.e., the posterior state distribution for a given action and observation, can be readily computed through a Bayesian update [48]:

$$b'(s') = b^{a,o}(s') = p(s'|o,a,b) = \frac{p(o|s',a)}{p(o|b,a)} \underbrace{\sum_{s \in S} p(s'|s,a)b(s)}_{b^a(s')} \quad (3)$$

where $p(o|b,a)$ is the standard normalizing constant, given as:

$$p(o|b,a) = \sum_{s' \in S} p(o|s',a) \sum_{s \in S} p(s'|s,a)b(s) \quad (4)$$

Following Eq. (3), beliefs can be updated as new actions are performed and new observations are collected, essentially encoding the information of the entire history of actions and observations up to the current time step $t$. As such, a new belief $b'$ is a sufficient statistic of the history of actions and observations up to $t$. Namely, by forming a belief about its state using Eq. (3), the agent has all the information required for deciding on an action. The policy in Eq. (2) can then be equivalently expressed as a mapping from beliefs to actions, $\pi : B \to A$.

It also follows from Eq. (3) that the agent moves from one belief to another based on the selected action and received observation. We can, thus, define the transition probability from belief $b$ to belief $b'$ as [54]:

$$p(b'|b,a) = \sum_{o \in \bar{\Omega}} p(o|b,a) \quad (5)$$

where $\bar{\Omega} \subseteq \Omega$ is the subset of observations leading to $b'$, when starting at belief $b$ and taking action $a$. Owing to Eq. (5), a POMDP can be seen as a belief-MDP, where transitions pertain to belief points, instead of states. For a given observation, which depends on the actual system state, the respective probabilistic graph is shown in Fig. 2. The belief-MDP reward $r_b = r_b(b,a)$ is the expected reward at the current step, which in the context of inspection and maintenance planning can be defined as [44]:

$$r_b(b,a) = \sum_{s \in S} b(s)r(s,a)$$
$$= \sum_{s \in S} b(s)(r_M + \gamma r_O + r_D)$$
$$= \underbrace{b \cdot R_{M,a}}_{r_{b,M} \text{ exp. maintenance cost}} + \underbrace{\gamma b \cdot R_{O,a}}_{r_{b,O} \text{ exp. observation cost}} + \underbrace{b \cdot R_D}_{r_{b,D} \text{ exp. damage cost}} \quad (6)$$

where reward $r$ (reward matrix $R$) is decomposed into $r_M$, $r_O$ and $r_D$ ($R_M$, $R_O$ and $R_D$), which are the maintenance action, observation action and damage state rewards (non-positive to reflect costs), respectively. Maintenance cost rewards pertain to interventions, such as retrofits, repairs, replacements, etc. Observation action rewards include the costs related to the type of data collection method and can, for example, refer to a visual inspection versus an ultrasonic inspection, or the installation of a monitoring system. Observations, $o$, are the outcomes of observation actions. An observation is assumed to convey information if it can change the posterior probability over the system states, as this is described by Eq. (3). An observation can thus be informative or uninformative, as defined below, and as also discussed in Section 3.

**Definition 1.** An *uninformative observation, $o \in \Omega$*, is an observation that does not change a prior belief, i.e., in a POMDP context, $b^{a,o}(s) = b^a(s)$, for all $s \in S$.

From Definition 1 and Eq. (3), it readily follows that if $\Omega$ is a unit set, i.e. $|\Omega|=1$, observations are always uninformative, for all $s \in S$ and $a \in A$.

Expected damage cost in Eq. (6) depends merely on the current state distribution (belief), and may be decomposed into more components pertaining for example to economic losses due to system downtime or shutdown, or costs related to various societal and environmental metrics (casualties, energy consumption, $CO_2$ equivalent emissions, among others, e.g. in [34]).

The expected return under any policy, $\pi$, defines the value function, $V^\pi$, whereas the expected return under the optimal policy defines the optimal value function, $V^*$. Exploiting the concept of belief-MDPs, we can use the Bellman equation [55], expressing the optimal value function as [48]:

$$V^*(b) = JV^*(b)$$
$$= \max_{a \in A} \left[ r_b(b,a) + \gamma \mathbb{E}_o \left[ V^*(b') \right] \right] \quad (7)$$
$$= \max_{a \in A} \left[ r_b(b,a) + \gamma \sum_{o \in \Omega} p(o|b,a) V^*(b') \right]$$

where $J$ is the Bellman operator for the belief-MDP problem defined by tuple $\mathcal{L}$, and $b'$ is the posterior state distribution at the next step given an action and an observation, as described by Eq. (3). $J$ is a contraction operator with fixed point $V*$ [37].

It should be noted that Eq. (7) is defined over the continuous space of the belief simplex, $B$, which essentially consists of an infinite number of beliefs. However, it has been proven that the optimal value function is piece-wise linear and convex, and can thus be described by a finite number of affine hyperplanes [56]. This important result reduces the decision problem to determining a finite set of vectors, also known as the $\alpha$-vectors:

$$V^*(b) = \max_{\hat{\alpha} \in \Gamma} \sum_{s \in S} b(s) \hat{\alpha}(s) \quad (8)$$

where $\Gamma$ is the set comprising all $\alpha$-vectors. Substituting Eqs. (4), (8) in Eq. (7) we obtain the detailed expression of the POMDP optimal value function:

$$V^*(\mathbf{b}) = \max_{a \in A} \Bigg[ \sum_{s \in S} b(s)(r_M + \gamma r_O + r_D) + \gamma \sum_{o \in \Omega} \max_{\mathbf{\alpha} \in \Gamma} \sum_{s \in S} b(s) \sum_{s' \in S} p(o|s',a) p(s'|s,a) \hat{\alpha}(s') \Bigg] \quad (9)$$

Eq. (9) can be solved using value iteration on the space of $\alpha$-vectors. However, performing exact value iteration on the vector space is generally impractical, except for small POMDP problems, since the new set of alpha vectors generated at every iteration step scales exponentially with the cardinality of the observation set, $|\Omega|$ [57].

## 2.1. Point-Based POMDP Algorithms

Point-based solvers adopt the concept of belief-MDPs and manage to alleviate the POMDP complexity by avoiding the exponential increase of $\alpha$-vectors. The idea is to restrict value iteration operations to a meaningful collection of discrete belief points, i.e., to perform $\alpha$-vector backups on a finite subset of the belief space, $\bar{B} = \{\mathbf{b}_0, \mathbf{b}_1, ...\} \subseteq B$, which is considered to be able to sufficiently approximate the original continuous ($|S|$-1)-dimensional simplex. Point-based algorithms take advantage of the fact that despite the continuity of the belief space, in practice there is only a finite number of belief points that are actually visited. These belief points lie in a reachable subset of the belief space, with respect to an initial (root) belief $\mathbf{b}_0$. At every iteration step, new $\alpha$-vectors are generated merely based on these points, forming a set $\bar{\Gamma} \subseteq \Gamma$ that can efficiently recover the true value function over the entire belief space, with the aid of the max operator of Eq. (8). Of course, since the $\alpha$-vectors in $\bar{\Gamma}$ cover the entire space, $B$, one can also compute an estimate of the value function for non-reachable beliefs, however, this estimate may be expected to be of lower accuracy. At each iteration, $\bar{\Gamma}$ is updated through every $\mathbf{b} \in \bar{B}$, or a subset of it, based on the *backup* operator defined as:

$$\text{backup}(\bar{\Gamma}, \mathbf{b}) = \arg\max_{\hat{\alpha}^a \in \bar{\Gamma}} \sum_{s \in S} b(s) \hat{\alpha}^a(s) \quad (10)$$

$$\hat{\alpha}^a(s) = r_M + \gamma r_O + r_D + \gamma \sum_{o \in \Omega} \sum_{s' \in S} p(o|s',a) p(s'|s,a) \hat{\alpha}^{a,o}(s') \quad (11)$$

$$\hat{\mathbf{\alpha}}^{a,o} = \arg\max_{\hat{\alpha} \in \bar{\Gamma}} \sum_{s \in S} b(s) \sum_{s' \in S} p(o|s',a) p(s'|s,a) \hat{\alpha}(s') \quad (12)$$

All point-based solvers maintain a lower bound on the value function, which is updated throughout the iteration process, as described in Eqs. (10)-(12), e.g. [57, 58, 59, 60]. This lower bound consists of the linear hyperplanes defined in Eq. (8), and is typically initiated by evaluating a simple policy. Modern point-based algorithms also compute, maintain and update an approximate upper bound on the value function. This bound allows these algorithms to employ more efficient strategies for belief space exploration, as well as to monitor convergence over the course of the iterative procedure. ZMDP with its Heuristic Value Iteration (HSVI) and Focused Real-Time Dynamic Programming (FRTDP) variants [59, 61], as well as Successive Approximation of the Reachable Space under Optimal Policies (SARSOP) [60] belong to this class of algorithms. The upper bound is typically initiated with optimistic values and, similarly to the lower bound, should be constructed as a piece-wise linear and convex function. However, it is not possible to update or evaluate the upper bound over the entire belief simplex using Eqs. (10)-(12), due to the presence of the max operator. Thus, the upper bound can be maintained by point-wise value estimates at visited beliefs and the formed convex hull that they support, which is determined through linear programming. Point-based solvers avoid solving this expensive linear program however and, instead, determine the upper bound using a much faster sawtooth approximation, since as the number of beliefs supporting the upper bound estimates increases, the linear program becomes considerably difficult to solve [37].

The points of $\bar{B}$ are either collected through randomly sampled belief trajectories, i.e., based on random sequences of actions and observations, or through more focused and informed search heuristics. The Point-Based Value Iteration (PBVI) algorithm [57], the first point-based algorithm, iterates between backup and belief space expansion steps. PBVI proposes an exploration strategy which expands over the existing points of $\bar{B}$, at every iteration. For every existing belief point, its successor is added to $\bar{B}$ such that the new set spreads as sparsely as possible over $B$. PBVI updates $\alpha$-vectors over all collected beliefs. The Perseus algorithm [58], traverses a series of path trials based on randomly sampled action and observation histories, in order to form $\bar{B}$, at the beginning of the solution procedure. This set of collected points remains unchanged during the $\alpha$-vector backups. Perseus also performs asynchronous randomized backups, i.e., it does not perform backups over all beliefs in $\bar{B}$, but instead selects randomly which belief values to update at every iteration step. Beliefs whose value is improved by $\alpha$-vectors supporting previously selected beliefs, are not updated in the current step. ZMDP and SARSOP utilize both the lower and upper bounds to inform the exploration of the belief space, choosing actions based on the upper bound and observations based on the maximum lower-upper bound gap. Both algorithms perform asynchronous bound updates over the visited beliefs.

In addition to their advanced exploration strategies, HSVI, FRTFP and SARSOP also apply pruning techniques to reduce the complexity and memory requirements related to the expansion of the $\alpha$-vectors set, removing vectors from $\bar{\Gamma}$ that are considered to be suboptimal under certain criteria. HSVI and FRTDP prune vectors that do not support at least one of the collected belief points and their immediate successors. The above algorithms can also optionally implement a *masking* technique which essentially tries to create compressed representations of the $\alpha$-vectors, by maintaining and updating $\alpha$-vector entries that are not zero or not close to zero. Similarly, SARSOP prunes vectors that either do not support at least one of the collected belief points or are dominated by other $\alpha$-vectors within a predefined neighborhood. SARSOP also prunes beliefs that are considered to be suboptimal based on the current information provided by the upper and lower bounds. Thereby, the entire tree of successors under these beliefs is pruned and exploration is restricted to more optimally reachable belief subspaces.

A detailed overview on point-based solvers along with their application in various robotic tasks can be found in [37]. Their insights and application details in structural inspection and maintenance planning can be found in [48, 46], where different point-based approaches are tested. Among them, the three most competitive are identified and used herein. Overall, it is demonstrated that point-based solvers can provide comprehensive and efficient near-optimal solutions in problems with thousands of states and a much lower number of actions and observations. In cases featuring more complex POMDP settings, deep reinforcement learning actor critic architectures have been shown to have significant success, as presented in [49]. The multi-agent actor critic approaches developed in [49, 51] combine belief-MDPs with decentralized deep reinforcement learning concepts and are able to learn detailed non-stationary life-cycle inspection and maintenance policies for engineering system settings with multiple components, operating in extremely large state, action and observation spaces.

## 3. Quantifying Value-based Information Gains

### 3.1. Step-wise Value of Information in POMDPs

As described above, a POMDP can be defined through a tuple $\mathcal{L} = \langle S, A, \mathbf{P}, \Omega, \mathbf{O}, \mathbf{R}, \gamma \rangle$. Based on the decomposable nature of the reward and the effects of different actions observational and intervention actions, the tuple can be re-written in a detailed form as $\mathcal{L} = \langle S, A_M \times A_O, [\mathbf{P}_{a_M}]_{a_M \in A_M}, \Omega_e \times \Omega_O, [\mathbf{O}_{a_O}]_{a_O \in A_O}, \mathbf{R}_M + \mathbf{R}_O + \mathbf{R}_D, \gamma \rangle$. $A_M$ is a set of maintenance actions $a_M$; $A_O$ is a set of observation actions $a_O$; $\mathbf{P}_{a_M}$ is the transition model for different maintenance actions $a_M$; $\Omega_e$ is a set of *default observations*; $\Omega_O$ is a set of observations, and is a union of observation sets $\Omega_{a_O}$ of the different observation actions $a_O$; $\mathbf{O}_{a_O}$ is the observation model for different observation actions $a_O$; $\mathbf{R}_M$, $\mathbf{R}_O$, $\mathbf{R}_D$ are the reward matrices as previously defined. Although for notational efficiency we assume the reward matrices to have the same dimensions $|S| \times |A|$, the maintenance costs are independent of $a_O$, the observation action costs are independent of $a_M$, whereas the damage costs are independent of both.

**Definition 2**. A *default observation*, $o_e \in \Omega_e$, is an observation which the decision-maker always receives from the environment, regardless of the selected action, i.e. $p(o_e | s, a) = p(o_e | s)$, for all $s \in S$ and $a \in A$.

**Definition 3**. A *trivial observation action*, $a_O \in A_O$, is an observation action with no cost, i.e. $r_O(s, a_O)=0$, for all $s \in S$ and $a_O \in A_O$, with its respective $\Omega_{a_O}$ being a unit set.

According to Definitions 1-3, if the decision-maker chooses a trivial observation action, it will receive the default observation from $\Omega_e$ plus an uninformative observation from $\Omega_O$. Thus, the decision-maker will overall receive the default observation, i.e. $\Omega \equiv \Omega_e$. Default observations are not necessarily uninformative, hence the respective POMDPs do not necessarily imply that the trivial actions yield no information. In deteriorating systems, failure or near-failure states, for example, are often self-announcing, meaning that they are "observable" regardless of the selected observation action.

Similarly, the trivial maintenance action is an action with no cost, i.e $r_M=0$, and yields a natural (uncontrolled) environment transition. As denoted by the respective subscripts, state transitions $\mathbf{P}_{a_M}$ merely depend on maintenance actions, meaning that only maintenance actions $a_M$ can change the state of the system, whereas observation actions $a_O$ can only change the agent's perception about the state of the system, thus perfectly sufficing to define the observation model $\mathbf{O}_{a_O}$. Based on the above, we can define the step-wise VoI associated with a certain policy $\pi$ as:

$$\text{VoI}_{step}^\pi (a_O) = \mathbb{E}_{o_e, o_O}\left[V^\pi\left(\mathbf{b}^{a_M, a_O, o_e, o_O}\right)\right] - \mathbb{E}_{o_e}\left[V^\pi\left(\mathbf{b}^{a_M, o_e}\right)\right] \quad (13)$$

Eq. (13) describes the gain the decision-maker expects when taking an observation action at a certain time step *t*, following a policy $\pi$ in the future. Subtracting the actual cost of the observation action from this gain, we obtain the net step VoI under a policy $\pi$ as:

$$\text{netVoI}_{step}^\pi (a_O) = \text{VoI}_{step}^\pi - |r_{b,O}| \quad (14)$$

Net step VoI expresses the net gain at step *t* as a result of additional information, also considering the cost to acquire this information (e.g. inspection cost). If nontrivial observation actions reveal the actual state of the system with certainty, i.e. $\mathbf{O}_{nontrivial}=\mathbf{I}$ (identity matrix), we can similarly define the step-wise Value of Perfect Information (step-wise VoPI), $\text{VoPI}_{step}^\pi$, and net step-wise VoPI, $\text{netVoPI}_{step}^\pi$, similarly to Eqs. (13) and (14). In such a case, in the term $\mathbb{E}_{o_e, o_O}\left[V^\pi\left(\mathbf{b}^{a_M, a_O, o_e, o_O}\right)\right]$ of Eq. (13), uncertainty is only attributed to the state transition, which is controlled by the chosen maintenance, $\text{VoPI}_{step}^\pi (a_O) = \mathbb{E}_{s' \sim \mathbf{b}^{a_M}}\left[V^\pi(s')\right] - \mathbb{E}_{o_e}\left[V^\pi\left(\mathbf{b}^{a_M, o_e}\right)\right]$.

**Lemma 1.** Any policy with convex value function on the belief simplex, *B*, has $\text{VoPI}_{step}^\pi \geq \text{VoI}_{step}^\pi \geq 0$.

*Proof*. Using basic probability definitions, Jensen's inequality, Eq. (3), and the fact that observation actions do not affect state transitions we can get:

$$\begin{aligned}
\mathbb{E}_{s' \sim \mathbf{b}^{a_M}}\left[V^\pi(s')\right] &= \mathbf{b}^{a_M} \cdot \left[V^\pi(s')\right]_{s' \in S} \\
&= \mathbb{E}_{o_e, o_O}\left[\mathbf{b}'\right] \cdot \left[V^\pi(s')\right]_{s' \in S} \\
&= \mathbb{E}_{o_e, o_O}\left[\mathbf{b}' \cdot \left[V^\pi(s')\right]_{s' \in S}\right] \\
&= \mathbb{E}_{o_e, o_O}\left[\mathbb{E}_{s' \sim \mathbf{b}'}\left[V^\pi\left(\left[\delta_{s'x}\right]_{x \in S}\right)\right]\right] \\
&\geq \mathbb{E}_{o_e, o_O}\left[\left[V^\pi\left(\left[\mathbb{E}_{s' \sim \mathbf{b}'}\left[\delta_{s'x}\right]\right]_{x \in S}\right)\right]\right] \\
&= \mathbb{E}_{o_e, o_O}\left[V^\pi\left(\mathbf{b}'\right)\right] = \mathbb{E}_{o_e, o_O}\left[V^\pi\left(\mathbf{b}^{a_M, a_O, o_e, o_O}\right)\right]
\end{aligned}$$

For the last expression, we further have:

$$\begin{aligned}
\mathbb{E}_{o_e, o_O}\left[V^\pi\left(\mathbf{b}^{a_M, a_O, o_e, o_O}\right)\right] &\geq \mathbb{E}_{o_e}\left[V^\pi\left(\mathbb{E}_{o_O}\left[\mathbf{b}^{a_M, a_O, o_e, o_O}\right]\right)\right] \\
&= \mathbb{E}_{o_e}\left[V^\pi\left(\mathbb{E}_{o_O}\left[\left[p(s | a_M, a_O, o_e, o_O, \mathbf{b})\right]_{s \in S}\right]\right)\right] \\
&= \mathbb{E}_{o_e}\left[V^\pi\left(\mathbf{b}^{a_M, a_O, o_e}\right)\right] \\
&= \mathbb{E}_{o_e}\left[V^\pi\left(\mathbf{b}^{a_M, o_e}\right)\right]
\end{aligned}$$

From the above, it immediately follows that inequalities

$\mathbb{E}_{s' \sim \mathbf{b}^{a_M}}\left[V^\pi(s')\right] \geq \mathbb{E}_{o_e,o_O}\left[V^\pi\left(\mathbf{b}^{a_M,a_O,o_e,o_O}\right)\right] \geq \mathbb{E}_{o_e}\left[V^\pi\left(\mathbf{b}^{a_M,o_e}\right)\right]$ hold, thus $\text{VoPI}_{step}^\pi \geq \text{VoI}_{step}^\pi \geq 0$. Equality $\text{VoI}_{step}^\pi = 0$ holds if $\Omega_{a_O}$ is a unit set, i.e. nontrivial observation actions also yield uninformative observations. Equality $\text{VoPI}_{step}^\pi = \text{VoI}_{step}^\pi$ holds if $\mathbf{O}_{nontrivial} = \mathbf{I}$, i.e. nontrivial observation actions reveal the actual system state with certainty. □

**Corollary 1**. Under the optimal POMDP policy, $\pi = \pi^*$, $\text{VoPI}_{step}^* \geq \text{VoI}_{step}^* \geq 0$.

*Proof.* The POMDP optimal value function is convex [56], so Lemma 1 holds. □

The above result can be also straightforwardly proven by using the specific piece-wise linear form of the optimal value function, $V^*(\mathbf{b}) = \max_{\hat{\alpha} \in \Gamma}\{\hat{\alpha} \cdot \mathbf{b}\}$, and the fact that $\sum \max(\cdot) \geq \max \sum(\cdot)$ [62]. Elaborating on Eq. (9), using Eqs. (13) and (14), we have:

$$V^*(\mathbf{b}) = \max_{a_M \in A_M} \left\{ r_{b,O^-} + \gamma \mathbb{E}_{o_e}\left[V\left(\mathbf{b}^{a_M,o_e}\right)\right] + \gamma \max_{a_O \in A_O}\left\{\text{netVoI}_{step}^*(a_O)\right\}\right\} \quad (15)$$

where $r_{b,O^-} = r_{b,M} + r_{b,D}$, i.e. combining any costs other than the expected inspection cost. The alternative statement of Bellman optimality in the belief space, as this is expressed by Eq. (15), is illustrative how the notion of information and its respective value is leveraged by a POMDP policy to guide inspections. Namely, for all possible maintenance actions, the decision-maker will take that observation action that maximizes the net VoI at this certain step.

**Corollary 2**. Under the optimal POMDP policy, $\pi = \pi^*$, if nontrivial observation actions are cost-free and informative then the decision-maker always observes.

*Proof.* Inspections are cost-free, i.e. $r_{b,O} = 0$ for all $a_O \in A_O$. Without loss of generality, we assume that $A_O = \{0,1,\ldots,|A_O|-1\}$, with $a_O=0$ denoting the trivial observation action. Then, using Eq. (14), (15) and Corollary 1 we obtain:

$$\arg\max_{a_O \in A_O}\left\{\text{netVoI}_{step}^*(a_O)\right\} = \arg\max_{a_O \in \{0,1,\ldots\}}\left\{\text{VoI}_{step}^*(0) - 0, \text{VoI}_{step}^*(1) - 0, \ldots\right\}$$

$$= \arg\max_{a_O \in \{0,1,\ldots\}}\left\{0, \underset{>0}{\text{VoI}_{step}^*(1)}, \ldots\right\} \neq 0 \quad □$$

**Corollary 3**. Under the optimal POMDP policy, $\pi = \pi^*$, $\max_{a_O \in A_O}\left\{\text{netVoI}_{step}^*\right\} \geq 0$.

*Proof.* Using Corollary 1, and Eq. (14), and assuming, without loss of generality, that $A_O = \{0,1,\ldots,|A_O|-1\}$, we can prove:

$$\max_{a_O \in A_O}\left\{\text{netVoI}_{step}^*(a_O)\right\} = \max_{a_O \in \{0,1,\ldots\}}\left\{\text{VoI}_{step}^*(0), \text{VoI}_{step}^*(1) - |r_{b,O}(1)|,\ldots\right\}$$

$$= \max_{a_O \in \{0,1,\ldots\}}\left\{0, \text{VoI}_{step}^*(1) - |r_{b,O}(1)|,\ldots\right\} \geq 0 \quad □$$

### 3.2. Life-Cycle Gain from Changing Control Setting

The expected life-cycle gain of one control setting versus another can be expressed as the value difference between the two settings, when different control action sets are available for each setting, but these apply to the same system, i.e., the two settings have the same state space and the same deterioration dynamics (transition model for the uncontrolled case), as well as the same discounted horizon. To quantify the value of expected life-cycle reward (or cost) of these two settings, we consider two tuples that define the following distinct POMDP problems:

$$\mathcal{L}_1 = \left\langle S, A_M^1 \times A_O^1, \left[\mathbf{P}_{a_M}^1\right]_{a_M \in A_M^1}, \Omega_e \times \Omega_O^1, \left[\mathbf{O}_{a_O}^1\right]_{a_O \in A_O^1}, \mathbf{R}_M^1 + \mathbf{R}_O^1 + \mathbf{R}_D, \gamma \right\rangle$$

$$\mathcal{L}_2 = \left\langle S, A_M^2 \times A_O^2, \left[\mathbf{P}_{a_M}^2\right]_{a_M \in A_M^2}, \Omega_e \times \Omega_O^2, \left[\mathbf{O}_{a_O}^2\right]_{a_O \in A_O^2}, \mathbf{R}_M^2 + \mathbf{R}_O^2 + \mathbf{R}_D, \gamma \right\rangle$$

(16)

Then, the expected life-cycle gain, $G_{\mathcal{L}_1,\mathcal{L}_2}$, from following the optimal policy in $\mathcal{L}_2$ versus $\mathcal{L}_1$, starting at any belief $\mathbf{b} \in B$, is computed as:

$$G_{\mathcal{L}_1,\mathcal{L}_2}(\mathbf{b}) = V_2^*(\mathbf{b}) - V_1^*(\mathbf{b}) \quad (17)$$

where $V_1^*, V_2^*$ are the optimal value functions of each tuple, $\mathcal{L}_1$, $\mathcal{L}_2$, respectively. Equivalently, Eq. (17) describes the expected benefit from changing a control scheme from $\mathcal{L}_1$ to $\mathcal{L}_2$ at belief $\mathbf{b}$.

To assess the expected life-cycle gain of one observational scheme versus another (e.g. SHM, inspection visits, etc.), the tuple elements related to maintenance actions have to be the same, thus one has to apply Eq. (17), considering the following POMDP problems:

$$\mathcal{L}_1 = \left\langle S, A_M \times A_O^1, \left[\mathbf{P}_{a_M}\right]_{a_M \in A_M}, \Omega_e \times \Omega_O^1, \left[\mathbf{O}_{a_O}^1\right]_{a_O \in A_O^1}, \mathbf{R}_M + \mathbf{R}_O^1 + \mathbf{R}_D, \gamma \right\rangle$$

$$\mathcal{L}_2 = \left\langle S, A_M \times A_O^2, \left[\mathbf{P}_{a_M}\right]_{a_M \in A_M}, \Omega_e \times \Omega_O^2, \left[\mathbf{O}_{a_O}^2\right]_{a_O \in A_O^2}, \mathbf{R}_M + \mathbf{R}_O^2 + \mathbf{R}_D, \gamma \right\rangle$$

(18)

For Eqs. (17), (18), $G_{\mathcal{L}_1,\mathcal{L}_2}$ is the expected life-cycle gain of two control settings which are merely discerned by their observation actions. In this case, Eq. (17) quantifies potential benefits as a result of different sources and/or accuracy of information. In the remainder of this section, we elaborate on special cases of Eqs. (17), (18) to derive the gains related to different observational schemes and their relation to VoI and VoSHM.

### 3.3. Value of Information

Considering Eq. (18), suppose $A_O^1$ is a unit set, containing only a trivial observation action. Then, $\mathbf{R}_O^1 = \mathbf{0}$. This technically means that $\Omega_O^1$ is defined by a unit set as well. As such, overall, from all states, only one observation is possible, which is the default observation, i.e. $\Omega^1 \equiv \Omega_e$. In this case, tuple $\mathcal{L}_1$ defines the *default* control problem (or otherwise often also called *prior* in the literature) of $\mathcal{L}_2$, i.e., $\mathcal{L}_1 \doteq \mathcal{L}_{def}$ and $\mathcal{L}_2 \doteq \mathcal{L}$, thus Eq. (17) gives the VoI of the observational scheme adopted in $\mathcal{L}_2$ [18]:

$$G_{\mathcal{L}_{def},\mathcal{L}}(\mathbf{b}) = VoI_{\mathcal{L}}(\mathbf{b}) = V^*(\mathbf{b}) - V^*_{def}(\mathbf{b}) \quad (19)$$

In addition to the previous assumption, let us now assume that $A_O^2$ is a unit set with only a nontrivial action available at no cost, and $|\Omega_O^2| = |S|$ with $\mathbf{O}^2_{nontrivial} = \mathbf{I}$ (identity matrix). In this case, the agent operates under perfect information at every decision step of $\mathcal{L}$. This reduces the POMDP defined by $\mathcal{L}$ to an MDP problem, i.e. $\mathcal{L} \doteq \mathcal{L}_{MDP}$. Under these assumptions, using Eq. (17) we obtain the Value of Perfect Information (VoPI):

$$G_{\mathcal{L}_{def},\mathcal{L}_{MDP}}(\mathbf{b}) = VoPI_{\mathcal{L}}(\mathbf{b}) = V^*_{MDP}(\mathbf{b}) - V^*_{def}(\mathbf{b}) \quad (20)$$

If the value functions in Eqs. (19) and (20) include the cost related to observational actions, then, according to [11], they can also be associated with the net VoI, as explained in the previous section. As intuitively understood and also formally proven below, VoPI is an upper bound of VoI, and both information gains should be non-negative, in the sense that information should not be expected to hurt decisions. Notwithstanding its intuitive nature, it is also showcased later that this remark is not necessarily true if the decision-maker is following an inspection and maintenance policy other than the optimal policy prescribed by the solution of Eq. (9). This is shown by counterexample in Section 4.1.3.

**Theorem 1**. Let $J_1$ and $J_2$ be two value function mappings defined on $\mathcal{V}_1$ and $\mathcal{V}_2$, such that:

- $J_1$ and $J_2$ are contractions with fixed points $V_1^*$ and $V_2^*$, respectively
- $V_1^* \in \mathcal{V}_2$ and $J_2 V_1^* \geq J_1 V_1^* = V_1^*$
- $J_2$ is an isotone mapping

Then $V_2^* \geq V_1^*$ is true.

*Proof.* See [62] page 87. □

**Proposition 1.** Under the optimal policies of the POMDPs defined by tuples $\mathcal{L}, \mathcal{L}_{MDP}, \mathcal{L}_{def}$, then $VoPI_{\mathcal{L}} \geq VoI_{\mathcal{L}} \geq 0$.

*Proof.* Following the results of Corollary 3, and Eq. (15), we obtain:

$$JV^*(\mathbf{b}) = \max_{a_M \in A_M}\left\{r_{b,O^-} + \gamma \mathbb{E}_{o_e}\left[V^*\left(\mathbf{b}^{a_M,o_e}\right)\right] + \gamma \max_{a_O \in A_O}\left\{\text{netVoI}^*_{step}(a_O)\right\}\right\}$$
$$\geq \max_{a_M \in A_M}\left\{r_{b,O^-} + \gamma \mathbb{E}_{o_e}\left[V^*\left(\mathbf{b}^{a_M,o_e}\right)\right]\right\}$$
$$= J_{def}V^*(\mathbf{b})$$

Following the results of Corollaries 1 and 2, and using Eq. (15) we obtain:

$$V^*(\mathbf{b}) = \max_{a_M \in A_M}\left\{r_{b,O^-} + \gamma \mathbb{E}_{o_e}\left[V\left(\mathbf{b}^{a_M,o_e}\right)\right] + \gamma \max_{a_O \in A_O}\left\{\text{netVoI}^*_{step}(a_O)\right\}\right\}$$
$$\leq \max_{a_M \in A_M}\left\{r_{b,O^-} + \gamma \mathbb{E}_{o_e}\left[V\left(\mathbf{b}^{a_M,o_e}\right)\right] + \gamma \max_{a_O \in A_O}\left\{\text{netVoPI}^*_{step}(a_O)\right\}\right\}$$
$$= \max_{a_M \in A_M}\left\{r_{b,O^-} + \gamma \mathbf{b}^{a_M} \cdot \left[V^{\pi}(s')\right]_{s' \in S}\right\}$$
$$= \mathbf{b} \cdot \max_{a_M \in A_M}\left\{\mathbf{R}_{M,a_M} + \mathbf{R}_D + \gamma \mathbf{P}_{a_M}^T \cdot \left[V^{\pi}(s')\right]_{s' \in S}\right\}$$
$$= J_{MDP}V^*(\mathbf{b})$$

$J$, $J_{def}$, $J_{MDP}$ are all contraction operators with fixed points $V^*, V^*_{MDP}, V^*_{def}$, i.e. the maximum expected discounted rewards (minimum cost) of the POMDPs defined by tuples $\mathcal{L}, \mathcal{L}_{MDP}, \mathcal{L}_{def}$. It can also be readily noticed that these operators describe isotone mappings, i.e. for $V_1 \geq V_2$ it holds that $JV_1 \geq JV_2$. Using Theorem 1, we then have that:

$$V^*_{def}(\mathbf{b}) \leq V^*(\mathbf{b}) \leq V^*_{MDP}(\mathbf{b})$$

It immediately follows that inequalities $VoPI_{\mathcal{L}} \geq VoI_{\mathcal{L}} \geq 0$ hold. □

**Proposition 2.** $VoI_{\mathcal{L}}$ and $VoPI_{\mathcal{L}}$ reach their highest values when default observations are always uninformative.

*Proof.* This result can be shown if one marginalizes out $o_e$ in Eq. (13), and similarly proceeds with the steps delineated in Lemma 1 and Proposition 1, for the step-wise and life-cycle metrics, respectively, noting that $\mathbb{E}_{o_e}\left[V^*\left(\mathbf{b}^{a_M,o_e}\right)\right] \geq V^*\left(\mathbf{b}^{a_M}\right)$. □

### 3.4. Value of Structural Health Monitoring

The VoSHM refers to the possible gains from investing in life-long SHM devices and practices, instead of, or in addition to, planning inspection visits at distinct times during the structural life-cycle. As such, the VoSHM relates to the critical decision, either at the design stage or later, of whether a monitoring scheme is worth to be adopted, and if so, of which type. VoSHM quantifies essentially the benefits of continuous data collection and information inflow in the decision-support system.

In this work, to quantify the VoSHM, we examine another special case of Eq. (18). We assume that $A_O^1$ contains at least one nontrivial available action. Conversely, $A_O^2$ contains only one available observation action which is, however, not the trivial one and is costless, i.e., $\mathbf{R}_O^2 = \mathbf{0}$. For the two POMDP settings, the nontrivial observation actions may follow different observation models. Thereby, $\mathcal{L}_1 \doteq \mathcal{L}_{1,opt}$ corresponds to the scenario of optional inspection visits, whereas $\mathcal{L}_2 \doteq \mathcal{L}_{2,perm}$ corresponds to an alternative observational scheme with permanent characteristics, as this provided by an SHM system. Along these lines, the VoSHM is defined as:

$$G_{\mathcal{L}_{1,opt},\mathcal{L}_{2,perm}}(\mathbf{b}) = VoSHM_{\mathcal{L}_{1,opt},\mathcal{L}_{2,perm}}(\mathbf{b}) = V^*_{2,perm}(\mathbf{b}) - V^*_{1,opt}(\mathbf{b}) \quad (21)$$

It should be noted that the expected VoSHM life-cycle gain defined in Eq. (21) cannot be strictly seen as VoI as it can also take negative values. This may happen, for example, if the state information provided by an optional inspection visit is more accurate than the outcome of the permanent monitoring system, for any possible reason. A VoSHM value lower than the cost of a SHM system (including acquirement, installation, maintenance, and operation costs, etc.) simply suggests that there is no benefit for the decision-maker to invest in SHM but, instead, optimal planning with selected inspection visits should be preferred.

Using Eq. (21) we can compute the VoSHM at every possible belief point that the system can visit throughout the planning horizon. Typically, the belief of foremost interest is the root belief, $\mathbf{b}_0$, which reflects the probability distribution over all possible states at the initial conditions, i.e., for the defined time step $t=0$. In this case, the VoSHM quantifies the life-cycle value of the monitoring system. For $t>0$, which usually corresponds to $\mathbf{b}_t \neq \mathbf{b}_0$, Eq. (21) describe the remaining VoSHM from that time onward. The notion of remaining VoSHM can be of particular practical importance in cases where the optimal salvage time of the SHM system needs to be determined.

If the nontrivial observation actions in $\mathcal{L}_{1,opt}$, $\mathcal{L}_{2,perm}$ share the same observation probability model, i.e., with the respective settings denoted as $\mathcal{L}_{1,opt} \doteq \mathcal{L}$, $\mathcal{L}_{2,perm} \doteq \mathcal{L}_{perm}$, we obtain a non-negative value in Eq. (21). This can technically refer to a case where both inspections and SHM are based on the same sensing units. Thus the VoSHM can be seen in this case as the Relative Value of Continuous Information (RVoCI), since it quantifies the possible gain if the nontrivial observation is continuously and freely available:

$$\begin{aligned} VoSHM_{\mathcal{L},\mathcal{L}_{perm}}(\mathbf{b}) &= RVoCI_{\mathcal{L}}(\mathbf{b}) \\ &= VoI_{\mathcal{L}_{perm}}(\mathbf{b}) - VoI_{\mathcal{L}}(\mathbf{b}) \\ &= V^*_{perm}(\mathbf{b}) - V^*(\mathbf{b}) \end{aligned} \quad (22)$$

**Proposition 3.** Under the optimal policies of the POMDPs defined by tuples $\mathcal{L}, \mathcal{L}_{perm}$, then $RVoCI_{\mathcal{L}} \geq 0$.

*Proof.* Using, Eq. (7) and Corollary 2 and the fact that $r_{b,O} \leq 0$ for all observation actions, we obtain:

$$\begin{aligned} JV^*(\mathbf{b}) &= \max_{a_M, a_O} \left\{ r_{b,O^-} + \gamma r_{b,O} + \gamma \mathbb{E}_{o_e,o_O}[V(\mathbf{b}')] \right\} \\ &\leq \max_{a_M, a_O} \left\{ r_{b,O^-} + \gamma \mathbb{E}_{o_e,o_O}[V^*(\mathbf{b}')] \right\} \\ &= \max_{a_M} \left\{ r_{b,O^-} + \gamma \mathbb{E}_{o_e,o_O}[V^*(\mathbf{b}')] \right\} \\ &= J_{perm}V^*(\mathbf{b}) \end{aligned}$$

As for Proposition 1, using Theorem 1, we have:

$$V^*(\mathbf{b}) \leq V^*_{perm}(\mathbf{b})$$

and it immediately follows that the inequality $RVoCI_{\mathcal{L}} \geq 0$ holds. □

## 4. Numerical Applications

We consider two inspection and maintenance problems and assess VoI, VoSHM and their specialized cases, for the underlying systems as discussed in Section 3. For both problems, the life-cycle analysis only includes the service life phase, without that of initial design and construction. The first problem pertains to a stationary three component system, whereas the second to a single-component structure, deteriorating according to a non-stationary corrosion model. For the reported results the point-based algorithms of FRTDP, SARSOP and Perseus have been implemented to solve the POMDP problems and to determine the optimal life-cycle strategies.

### 4.1. Three-Component Deteriorating System

#### 4.1.1. Environment and description of control settings

For the purposes of a parametric numerical investigation in the presence of various observability accuracy levels, we consider a small three-component system. An infinite horizon case with $\gamma=0.95$ is analyzed. The discount factor, $\gamma$, reflects the current value of future costs, thus largely depending on economic features, such as interest rate and inflation. In management of deteriorating infrastructure systems, annual values of discount factors typically range between 0.95-0.98 [39, 42, 34]. Stochastic deterioration of the components, for all $i \in \{1,2,3\}$, is defined by independent transition matrices, $\mathbf{P}_{(i),0}$, whereas whenever a repair action is taken the components share the same transition matrix $\mathbf{P}_{(i=1,2,3),rep}$:

$$\mathbf{P}_{(1),0} = \begin{bmatrix} 0.82 & 0.13 & 0.05 \\ & 0.87 & 0.13 \\ & & 1.00 \end{bmatrix}, \quad \mathbf{P}_{(2),0} = \begin{bmatrix} 0.72 & 0.19 & 0.09 \\ & 0.78 & 0.22 \\ & & 1.00 \end{bmatrix},$$

$$\mathbf{P}_{(3),0} = \begin{bmatrix} 0.79 & 0.17 & 0.04 \\ & 0.85 & 0.15 \\ & & 1.00 \end{bmatrix}, \mathbf{P}_{(i=1,2,3),rep} = \begin{bmatrix} 0.90 & 0.10 \\ 0.80 & 0.20 \\ 0.70 & 0.30 \end{bmatrix} \quad (23)$$

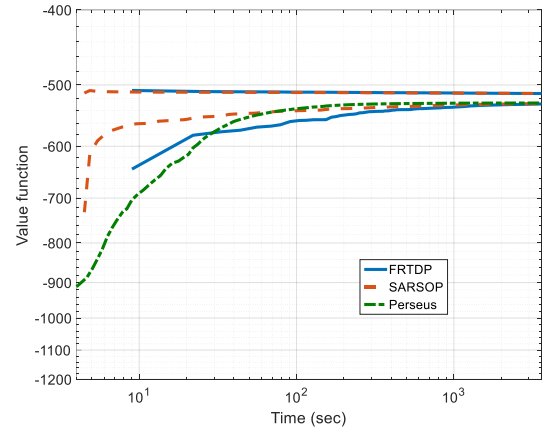

**Fig. 3**: Performance of different point-based POMDP algorithms in the three-component system problem, with $p=0.90$, for Setting 1 (optional monitoring).

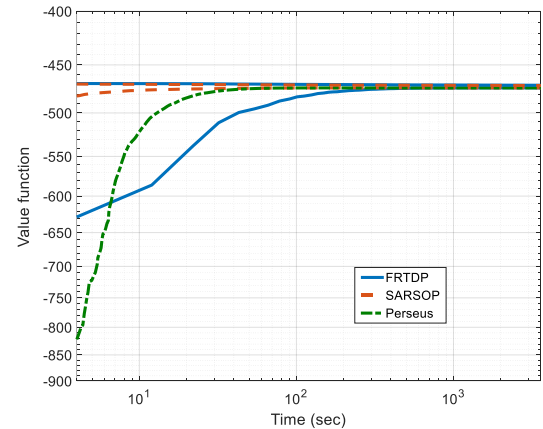

**Fig. 4**: Performance of different point-based POMDP algorithms in the three-component system problem, with $p=0.90$, for Setting 2 (permanent monitoring).

**Table 1**
Individual component costs (negative rewards) of maintenance and observation actions for three-component deteriorating system POMDP.

| Condition levels | | 1 | 2 | 3 |
|---|---|---|---|---|
| Maintenance rewards ($r_M$) | 1: Do nothing | 0 | 0 | 0 |
| | 2: Repair | -12 | -18 | -30 |
| Observation rewards ($r_O$) | 1: No observation | 0 | 0 | 0 |
| | 2: Observation | -1 | -1 | -1 |
| Damage rewards ($r_D$) | | 0 | -5 | -12 |

As indicated by Eq. (23), each component is described by three condition levels with stationary transition dynamics, i.e., transition from condition level *k* to *j* is independent of time, component age or deterioration rate. For example, for component 3, the transition probability from state 1 to state 3 is 0.04. Overall, the examined system can be fully specified by 27 states. Markovian transition probabilities of structural systems can be constructed based on simulated or real data of longitudinal responses, system conditions, rankings, etc., e.g. in [38, 63, 64], either through maximum likelihood estimation, or expectation-maximization schemes in the presence of latent state variables.

In order to quantify the VoSHM for this three-component system, two POMDP control settings are evaluated. For Setting 1, 4 observation and maintenance control actions are available for each component, including the possibility of inspection visits at belief points suggested by the POMDP solution. These actions are *'no observation and no repair'*, *'observation and no repair'*, *'no observation and repair'*, and *'observation and repair'*. The *'no observation'* observation action is the trivial observation action, and the default observation is considered uninformative. As such, the default control problem is here called *blind*, $\mathcal{L}_{def} \doteq \mathcal{L}_{blind}$. The total number of system actions is 64. For Setting 2, observations of nontrivial actions are available at no cost at every decision step, corresponding to a permanent monitoring observational scheme. Accordingly, only 2 maintenance control actions need to be considered, i.e., *'no-repair'*, and *'repair'*. Based on the possible action combinations, 8 system actions are available for Setting 2. Observation matrices, for all components, are given as:

$$\mathbf{O}^{(i=1,2,3)} = \begin{bmatrix} p & (1-p)/2 & (1-p)/2 \\ (1-p)/2 & p & (1-p)/2 \\ (1-p)/2 & (1-p)/2 & p \end{bmatrix} \quad (24)$$

Eq. (24) assigns an observation accuracy of $0 \leq p \leq 1$ every time an *'observation'* is taken, meaning that the correct state is observed with probability *p*, whereas either one of the other states is observed uniformly at random with probability 1-*p*. Negative rewards (or costs) for individual components are given in Table 1 for different states and actions. Observation actions are considered to cost 1/12, 1/18, and 1/30 of the repair cost for condition levels 1,2,3 respectively. Observation actions have constant costs with respect to states, whereas repair costs are considered to increase with damage severity. These values establish representative proportions between inspection and repair costs [42, 27], and can vary as per the specific nature of the studied engineering system. System level interdependence among components is established though the reward function, with certain penalties added to the cumulative component costs at different system state configurations. That is, for system states {(2,2,1)}, {(2,2,2),(1,2,3),(2,2,3)}, {(3,3,1),(3,3,2)}, and {(3,3,3)}, penalties are -5.0, -10.0, -14.0, and -18.0, respectively, where vector (*i,j,k*) denotes component condition level combinations, i.e., (3,3,1) indicates that there are 2 components in condition level 3 and one component in condition level 1. These system-level state rewards are combined with the rewards of the individual components, shown in Table 1.

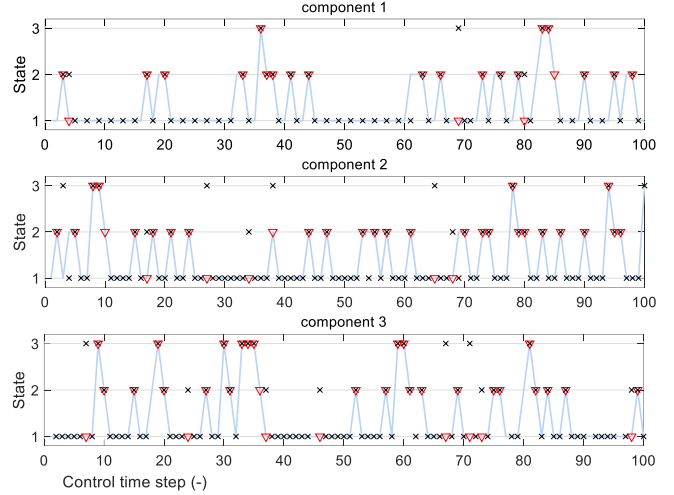

**Fig. 5**: Policy realization of three-component system Setting 1 (optional inspection), with *p*=0.90, for all components.

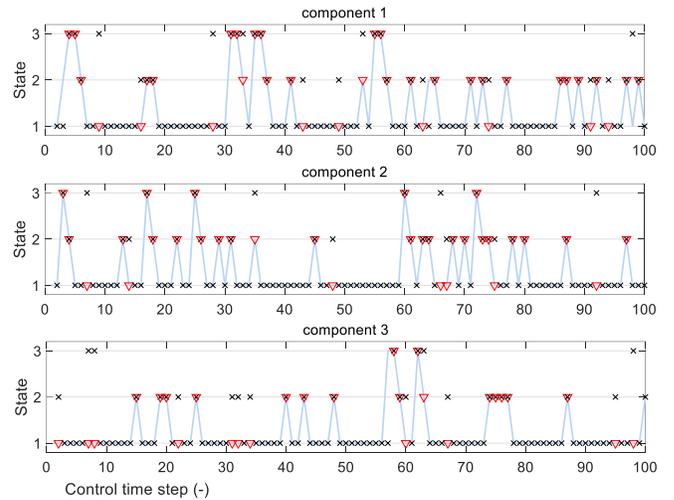

**Fig. 6**: Policy realization of three-component system Setting 2 (permanent monitoring), with *p*=0.90, for all components.

### 4.1.2. *Evaluation of optimal policies*

For both POMDP settings, FRTDP, SARSOP and Perseus point-based algorithms are implemented. As shown in the analysis results presented in Figs. 3 and 4, for $p$=0.90, Setting 1 practically converges after 1,000s, whereas Setting 2 after 110s for all algorithms. It can be seen that the precision of the solution of Setting 1 is somewhat lower that the precision of Setting 2, for FRTDP and SARSOP. This can be attributed to the fact that the system in Setting 1 operates in a much more challenging POMDP environment with more actions and, consequently, larger reachable belief space. Apart from that, low precision can also be triggered by a rough approximation of the upper bound. As discussed in Section 2.1, FRTDP and SARSOP utilize approximate upper bounds, determined by a sawtooth approximation. The bound that actually contains all the information of the optimal policy is the lower bound and this is shown to be reached with great agreement among the different algorithms. Overall, in Fig. 3 SARSOP converges faster, thus exhibiting a better anytime performance, as also discussed in [48]. Perseus, although starting from a cruder initial lower bound, eventually reaches the best value, slightly outperforming its counterparts. The same features are also noticed in Fig. 4, where the overall convergence is much faster for all algorithms, due to the simpler nature of the decision problem. SARSOP demonstrates considerable strengths in early convergence, practically converging before 10s. Perseus has an anytime performance advantage compared to FRTDP, whereas all solvers reach identical lower bounds after 3,600s.

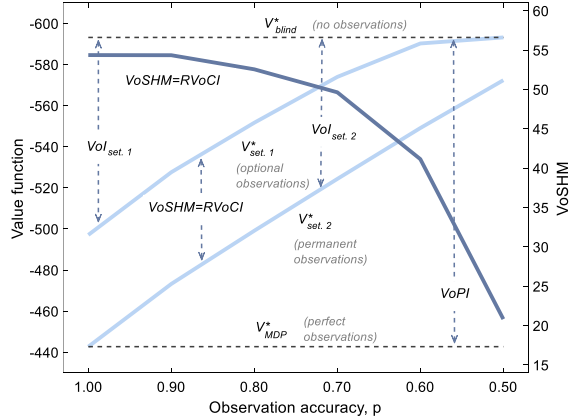

**Fig. 7**: Optimal value functions of three-component system settings 1 and 2 and respective VoSHM, for different observability levels.

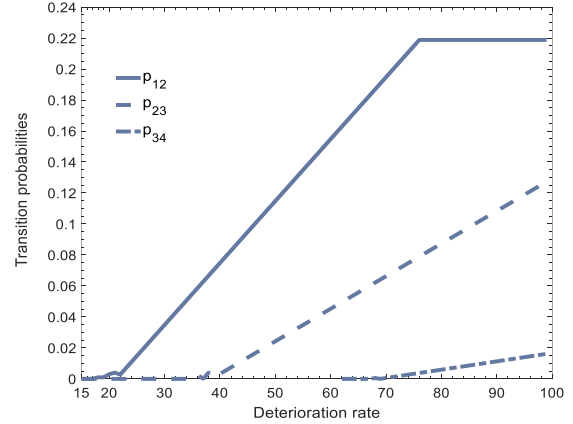

**Fig. 8**: Transition probabilities between adjacent structural conditions as functions of the deterioration rate.

A realization of the converged policy is shown in Figs. 5 and 6. For Settings 1 and 2, each component needs to perform different policies in order for their combined behavior to collectively minimize the total expected cost of the system. In Fig. 5, depicting a policy realization for the case of optional inspections, component 1 requires an inspection visit roughly every two years, whereas its *'repair'* actions are mostly taken at the inspection times. Component 2 requires inspections at almost every decision step (all time steps except $t$=10 in the realization of Fig. 5. Component 3 policy combines features of the other two policies, choosing frequent inspections, with a few *'no observation and no repair'* actions. These policy patterns are intuitively anticipated as the transition dynamics of component 3 are in-between the other two cases defined by components 1, 2. Fig. 6 illustrates a life-cycle policy realization for the case of permanent monitoring (Setting 2). In this POMDP setting, observations are always available at no cost due to the permanent monitoring system assumption, as explained in Section 3.2.

The converged value functions and VoI for each setting, as well as the VoSHM are shown as functions of the observability accuracy level, $p$, in Fig. 7. VoSHM equals the RVoCI, as Settings 1 and 2 share the same observation matrices for their observation actions. It can be observed that as the observation accuracy increases, the VoSHM increases and is concave down, reaching a plateau at higher levels of accuracy. The VoSHM of the system ranges from ~3% to ~11% of the value of Setting 1, for $p$=0.50 to $p$=1.00, respectively. This means that any permanent monitoring system with lifetime cost lower than these amounts should be preferred, in place of any inspection plan, including the optimal one. The VoI also increases with increased observability, for both settings, however it is concave up. This pattern is more prominent for the value function of Setting 1, where a plateau is practically reached for $p$<0.60. This indicates that the observation quality is quite poor at this region, so the decision-maker does not choose to pay for inspection and, consequently, the value of Setting 1 becomes equal to the value of the optimal blind policy. The VoPI is ~25% of the optimal blind policy cost and, by definition, is reached by the VoI of Setting 2, for $p$=1.00.

*4.1.3. Can better information hurt?*

As briefly discussed and proven in Sections 3.3 and 3.4, VoI, VoPI and RVoCI describe non-negative gains under the optimal policy provided by Eq. (7). This practically implies that if the decision-maker follows the optimal POMDP policy, which is also the globally optimal policy as long as the problem adheres to the dynamic programming principle of optimality, there is no possibility that more and/or better information can lead to worse decisions, thus to a higher life-cycle cost. However, this fundamental, intuitive property that "information never hurts" does not necessarily hold true for policies that are only locally optimal at certain subsets of the policy space, or otherwise suboptimal.

To further illustrate this remark, we again consider the same deteriorating system examined in this section and we now focus, out of all possible policies, on the locally optimal solutions corresponding to the policy subspace of some condition-based maintenance policies. Accordingly, repairs are now decided based on the condition observation outcomes (and not the belief), and the same condition-repair pairs are optimized for all components. We further consider that the default observation of the environment follows the observation model of Eq. (24) with $p=0.96$, and there is also a cost-free nontrivial observation following the same model with $p=1.0$. Thus, $p=0.96$ characterizes the default control setting, whereas $p=1.0$ characterizes a perfectly observable setting. Note that in this case VoI≡VoPI, since the nontrivial observation action will be always chosen in the latter setting. In both settings, the optimal condition-based maintenance policies are "repair if state 3 is observed, do nothing otherwise". After quantifying the relevant optimal condition-based maintenance policies for the two scenarios, it is found that the life-cycle cost of the default setting is 665.09, and that of the perfectly observable setting is 665.94, with their 99% confidence intervals in the order of 0.22. VoI and VoPI are thus negative here. This showcases that better information does possibly hurt, if it is not integrated within an efficient decision optimization framework.

### 4.2. Corroding Deck Structure

#### 4.2.1. Environment and description of control settings

The reinforced concrete port structure under corrosion presented in is studied in this example. The original deteriorating environment for this system is described by 4 condition levels. The non-Markovian characteristics of this corroding environment are addressed by combining the 4 condition levels with 83 corrosion rates [38]. The discount rate for this problem is set $\gamma=0.95$. The Markovian deterioration of the system, corresponding to the uncontrolled system evolution, is computed by a physically-based corrosion model [65], and is of the following form:

$$\mathbf{P}_0 = \left[ p(x_{t+1}=j, \tau_{t+1}=\tau+1 | x_t=i, \tau_t=\tau) \right]_{\substack{i,j\in\{1,\ldots,4\} \\ \tau\in\{1,\ldots,83\}}}$$

$$= \begin{bmatrix} p_{\tau,11} & p_{\tau,12} & & \\ & p_{\tau,22} & p_{\tau,23} & \\ & & p_{\tau,33} & p_{\tau,34} \\ & & & p_{\tau,44} \end{bmatrix} \quad (25)$$

where $x$ is the condition level and $\tau$ is the deterioration rate. The values of the probabilities in Eq. (25) are shown in Fig. 8. To account for finite horizon policies we appropriately augment the state space with respect to different time steps, thus finally the structure is defined by 14,009 states in total. There are 4 available actions related to maintenance interventions or replacements, namely *'no repair'*,

**Table 2**
Costs (negative rewards) of maintenance and observation actions for corroding deck structure POMDP.

| Condition levels | | 1 | 2 | 3 | 4 |
|---|---|---|---|---|---|
| | 1: Do nothing | 0 | 0 | 0 | 0 |
| Maintenance rewards ($r_M$) | 2: Repair | -60 | -110 | -160 | -280 |
| | 3: Major repair | -105 | -195 | -290 | -390 |
| | 4: Replace | -820 | -820 | -820 | -820 |
| | 1: No observation | 0 | 0 | 0 | 0 |
| Inspection rewards ($r_O$) | 2: Visual observ. | -4.5 | -4.5 | -4.5 | -4.5 |
| | 3: Monit. observ. | -7.5 | -7.5 | -7.5 | -7.5 |
| Damage rewards ($r_D$) | | -5 | -40 | -120 | -250 |

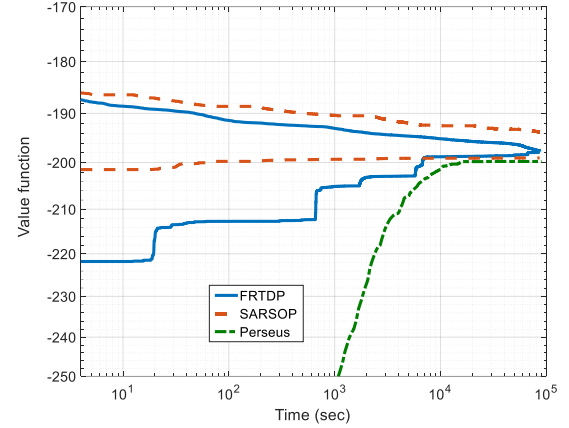

**Fig. 9**: Performance of different point-based POMDP algorithms in the corroding concrete port deck structure problem, for Setting 1 (optional inspection).

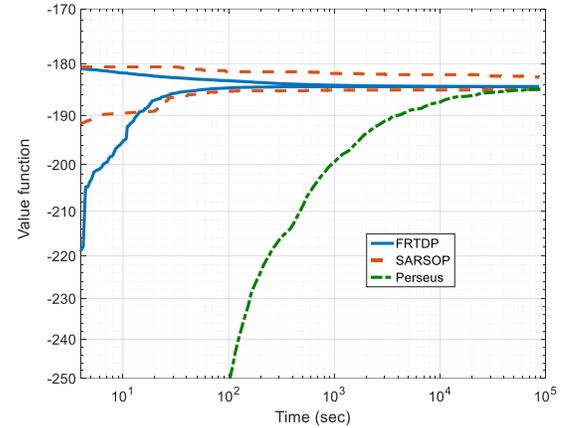

**Fig. 10**: Performance of different point-based POMDP algorithms in the corroding concrete port deck structure problem, for Setting 2 (permanent monitoring).

*'minor repair'*, *'major repair'* and *'replace'*. The transition for the *'no repair'* action follows Eq. (25). The *'minor repair'* action influences only the condition level transition of the system, whereas the *'major repair'* action influences both the condition level and the deterioration rate transition (deterioration rate is reduced by 3 steps). The full transition probability matrices for all maintenance actions can be found in [38].

For Setting 1 (optional inspections), the decision maker has 3 available inspection actions, *'no observation'*, *'visual observation'* and *'monitoring observation'*. For the complete action set, including the maintenance and the observations, we form 10 actions, instead of 12, since due to the nature of transition probabilities of the *'replace'* action, possible replacements do not need to be combined with observations. The observation matrices for the two nontrivial observations are:

$$\mathbf{O}_{vis.} = \begin{bmatrix} 0.63 & 0.37 & & \\ 0.10 & 0.63 & 0.27 & \\ & 0.10 & 0.63 & 0.27 \\ & & 0.20 & 0.80 \end{bmatrix} \quad (26)$$

**Table 3**
Life-cycle cost (negative reward) estimates with 95% confidence bounds and corresponding value of structural health monitoring, for three point-based algorithms.

| Algorithm | Setting 1 | Setting 2 | VoSHM |
|---|---|---|---|
| FRTDP | -198.253 ± 1.042 | -181.126 ± 3.830 | 17.127 ± 4.872 |
| SARSOP | -198.549 ± 2.512 | -184.437 ± 2.187 | 14.112 ± 4.699 |
| Perseus | -199.015 ± 2.829 | -183.043 ± 2.168 | 15.972 ± 4.997 |

$$\mathbf{O}_{mon.} = \begin{bmatrix} 0.80 & 0.20 & & \\ 0.05 & 0.80 & 0.15 & \\ & 0.05 & 0.80 & 0.15 \\ & & 0.10 & 0.90 \end{bmatrix} \quad (27)$$

The values of the observation matrices reflect the probability (likelihood) of receiving an observation (columns), given a state (rows). It should be noted that the number of observations is not necessarily equal to the number of states. The typically used probability of detection (PoD) for example, e.g. [26, 27], can be given by an *n*-by-2 observation matrix, where *n* is the number of states and 2 is the number of observations, e.g. defect detection or not. The relevant negative rewards (costs) are shown in Table 2. For a detailed presentation of costs and other model assumptions the interested reader is referred to [44, 38]. For Setting 2 we consider a permanent observational scheme, which is assumed to capture the flow of information provided by an SHM system. The respective observation matrix is also described by Eq. (27). As discussed in Section 3.4, this is a default observation at no cost for the purposes of evaluating the VoSHM. As such, the assumption is adopted herein, for the purposes of this illustrative example, that the outcome of a non-destructive evaluation inspection has the same state updating effect as the SHM system outcome. This can also technically refer to a case where both inspections and SHM share the same type of sensors. For example, for this particular case of corroding reinforced concrete structure, relevant electrochemical sensing units can be either mobile (operated by an inspector) or permanently installed and embedded in the concrete.

*4.2.2. Evaluation of optimal policies*

The analysis results during the value iteration are shown for Settings 1 and 2 in Figs. 9 and 10, respectively. In Fig. 9, where the optional inspection setting is considered, we can observe that SARSOP has very good early performance, however FRTDP eventually converges after about 24h. This can be attributed to the masking technique of FRTDP which exploits the sparsity of *α*-vectors, thus accommodating better a sparse environment like the one considered in this example. Fig. 10, shows the convergence of the point-based solvers for the permanent monitoring case (Setting 2). Similar performance is observed regarding the solvers comparison, however, a very good near-optimal solution is discovered much faster. Indicatively, FRTDP converges after about 300s. Fast convergence in Setting 2 is anticipated, since the problem comprises only 4 actions, compared to 10 in Setting 1, so the overall reachable belief space is less extensive.

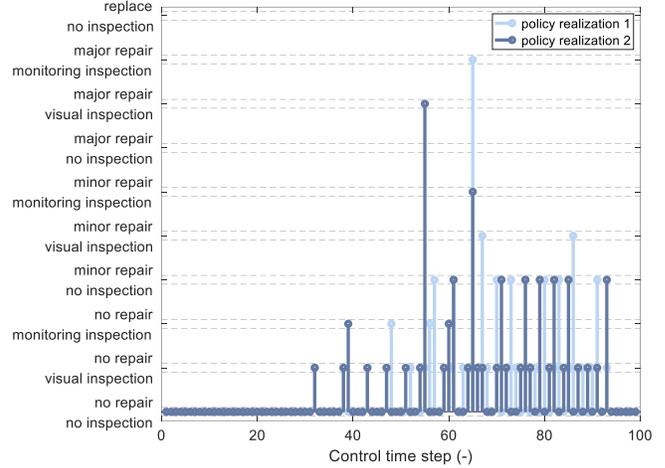

**Fig. 11**: Two policy realizations of corroding deck structure, for Setting 1 (optional inspection).

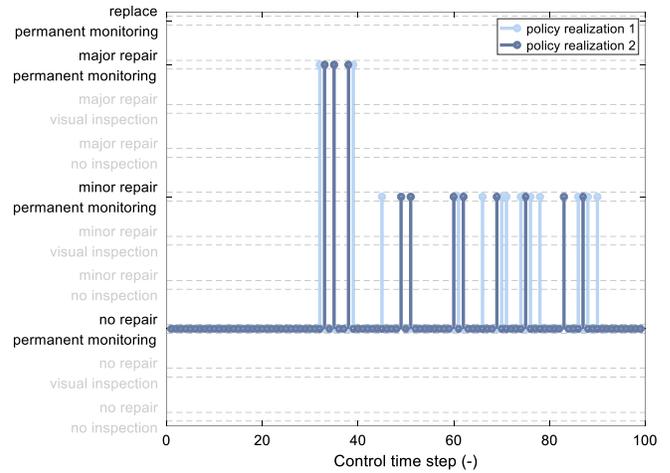

**Fig. 12**: Two policy realizations of corroding deck structure, for Setting 2 (permanent monitoring).

Using the converged lower bound, we also show two realizations of the optimal policies of Settings 1 and 2, in Figs. 11 and 12, respectively. In both Settings 1 and 2, the decision-maker starts taking nontrivial maintenance and observation actions after ~30 time steps. This happens because severe deterioration has not typically started until these time steps. In Setting 1, the agent mostly takes *'visual observation'* actions and a few *'monitoring observation'* actions, which are combined with *'no repair'* actions. Regarding maintenance actions, *'minor repair'* actions are shown to suffice for optimal control, along with a sparse selection of *'major repair'* actions. In Setting 2, the agent has a better understanding of its state, due to the presence of the permanent monitoring system. This gives the agent the opportunity to avoid taking any maintenance actions unless it is necessary due to expectation of high cost states. When this time comes, after ~30 time steps, it starts with *'major repair'* actions and then it proceeds with *'minor repairs'*, until stopping taking actions, after ~90 time steps. At the final steps of the realizations, in both settings, no maintenance and observation actions are selected. This happens as a result of the finite horizon policy, which means that the agent knows its exact time step, in addition to the belief over condition levels and the exact deterioration rate, before taking an action. As such, when approaching the final zero-valued absorbing state, which signifies end of the planning horizon, future cumulative state costs start becoming less significant, so no state corrections or better understanding of the system condition is required.

The results of the life-cycle cost estimates based on the policy described by the lower bound of the converged value functions, along with the respective VoSHM estimates, are shown in Table 3. Results of all the utilized point-based POMDP algorithms are shown with a maximum analysis time of 24h. It can be seen that the VoSHM is in the order of ~7% of the life-cycle cost estimate of Setting 1. As mentioned in the example of Section 4.1, this amount indicates the maximum cost the decision-maker should plan invest at the beginning of the control horizon, in order to acquire, install, operate and maintain a SHM system.

## 5. Conclusions

A methodology for quantifying the Value of Information (VoI) and the Value of Structural Health Monitoring (VoSHM) is presented in this work. The two metrics are defined over the life-cycle of the system, quantifying the expected life-cycle gains upon availability of inspection or monitoring information. A step-wise definition of VoI is also introduced and studied, showcasing that the notion of VoI is naturally utilized for selection of observational actions in POMDPs. POMDPs are employed to handle the overall optimal decision-making problem, including maintenance actions, which need to be scheduled together with observation actions for the assessment of life-cycle costs. Relevant theoretical analysis based on the above value-based definitions of the various information gains shows that, if the optimal POMDP policy is followed, information is optimally incorporated in the decision-making process and can only benefit the life-cycle cost. Accordingly, VoI and the Relevant Value of Continuous Information (RVoCI), introduced as a special case and proxy of VoSHM, are shown to be non-negative quantities, whereas both are upper bounded by their respective values when information is perfect. The above properties are notably found to not necessarily hold true for locally optimal or otherwise suboptimal policies. Optimal POMDP solutions are derived with the aid of point-based algorithms that can efficiently explore and evaluate the reachable belief space, from an initial system state distribution. Based on the above, VoI and VoSHM estimates are obtained based on pairs of different POMDP settings. These settings share the same state space, with the same stochastic deterioration properties, and operate over the same discounted horizon, having identical sets of maintenance actions. For the quantification of VoI, the first setting involves optimization of maintenance actions for the default problem (no observation actions available), whereas the second setting optimizes both maintenance and observation actions. For the quantification of VoSHM, the first setting corresponds to an observational scheme with optimal inspections, whereas the second setting operates under the assumption of continuous observations throughout the entire operational life, thus representing a permanent monitoring system. The results presented in a three-component deteriorating system and a single-component structure under corrosion indicate that the proposed approach provides a straightforward way to quantify the expected gains of different observational alternatives. The outcome of this analysis is a quantitative answer to the question of how much is information from inspections and/or monitoring worth, as well as how information of increased precision can affect decisions. Potential important extensions of the present work include, among others, consideration of different types of inspection and monitoring observation models directly calibrated based on real data, integration of advanced learning techniques with the decision-making process for online extraction of efficient damage and condition indicators from high-dimensional and heterogeneous SHM data, as well as VoSHM utilization for design of SHM systems and sensor placement.

## Acknowledgements

This material is based upon work supported by the U.S. National Science Foundation under CAREER Grant No. 1751941, and by CIAMTIS, the 2018 U.S. DOT Region 3 University Transportation Center. The authors would like to thank Hongda Gao for his early assistance on the first numerical application, during his stay at Penn State as a visiting doctoral student from the Beijing Institute of Technology. Prof. Chatzi would further like to acknowledge the support of the ERC Starting Grant WINDMIL on the topic of "Smart Monitoring, Inspection and Life-Cycle Assessment of Wind Turbines".